\documentclass[11pt,aps]{article}
\usepackage[sfdefault]{carlito} 
\usepackage{geometry}
\usepackage{setspace}
\usepackage{titlesec}
\PassOptionsToPackage{hyphens}{url}\usepackage{hyperref}
\usepackage{graphicx}
\usepackage{amsmath}
\usepackage{bm}
\usepackage{wrapfig}
\usepackage{xcolor}
\usepackage{authblk}
\usepackage{subcaption}
\usepackage{tikz}
\usetikzlibrary{matrix}
\usepackage[numbers,sort&compress]{natbib}
\usepackage{upgreek}

\titleformat{\section}{\normalfont\fontsize{11}{13}\bfseries\sffamily}{\thesection}{1em}{}
\titleformat{\subsection}{\normalfont\fontsize{11}{13}\bfseries\sffamily}{\thesubsection}{1em}{}
\titleformat{\title}{\normalfont\fontsize{14}{16}\bfseries\sffamily}{}{0em}{}

\title{Neural Network Modeling of Microstructure Complexity Using Digital Libraries}

\author[1]{Yingjie Zhao}
\author[1]{Zhiping Xu$^\ast$}
\affil[1]{Applied Mechanics Laboratory, Department of Engineering Mechanics, Tsinghua University, Beijing 100084, China}
\affil[$^\ast$]{Corresponding author(s): Zhiping Xu (xuzp@tsinghua.edu.cn)}

\date{}

\newlength{\tempdima}
\newcommand{\rowname}[1]
{\rotatebox{90}{\makebox[\tempdima][c]{\textbf{#1}}}}

\newcommand\zfig[1]{{\color{violet}#1}}

\usepackage[T1]{fontenc}
\usepackage{array}
\usepackage{booktabs}
\usepackage{float}

\usepackage{url}
\begin{document}
\maketitle

\begin{abstract}
Microstructure evolution in matter is often modeled numerically using field or level-set solvers, mirroring the dual representation of spatiotemporal complexity in terms of pixel or voxel data, and geometrical forms in vector graphics.
Motivated by this analog, as well as the structural and event-driven nature of artificial and spiking neural networks, respectively, we evaluate their performance in learning and predicting fatigue crack growth and Turing pattern development.
Predictions are made based on digital libraries constructed from computer simulations, which can be replaced by experimental data to lift the mathematical overconstraints of physics.
Our assessment suggests that the leaky integrate-and-fire neuron model offers superior predictive accuracy with fewer parameters and less memory usage, alleviating the accuracy-cost tradeoff in contrast to the common practices in computer vision tasks.
Examination of network architectures shows that these benefits arise from its reduced weight range and sparser connections.
The study highlights the capability of event-driven models in tackling problems with evolutionary bulk-phase and interface behaviors using the digital library approach.
\end{abstract}

%
\clearpage
\newpage

\section{Introduction}
The spatiotemporal evolution of microstructures is ubiquitous in nature, and often of vital importance in science and engineering (\zfig{Fig. 1a, b}).
Resolving this complexity is central to understanding and designing matter, which is, however, often difficult in practice due to the non-equilibrium and multiscale characteristics that cannot be fully captured by simplified theoretical models~\cite{chen2002ARMR}.
In addition to the bulk behaviors represented by field variables, interfaces also play an essential role in microstructure evolution, sometimes even more critically, because they facilitate key processes such as reaction, diffusion, mechanical degradation, and energy (e.g., thermal, electrical) transport.
Collecting experimental data at full resolution of the processes remains a challenge.
On the other hand, numerical frameworks such as level-set (LS, representing a sharp interface) and phase-field (PF, denoting a diffuse interface) methods were developed to track the evolutionary process.
The LS method utilizes a level set function to define the interface, where the zero level of this function represents the location of the interfaces~\cite{osher1988LS}.
PF uses continuous field variables to represent material phases and boundaries, removing the need to explicitly track interfaces and avoiding numerical singularities~\cite{cahn1958PF}.
While LS excels in simulating interface evolution, PF integrates coherently into multiphysics modeling frameworks~\cite{gaston2009moose}.

However, practical modeling of microstructure evolution is constrained by theoretical assumptions and uncertainties in parameterization~\cite{greene2013CMAME}.
The governing physics can be mathematically represented by partial differential equations (PDEs) with boundary conditions (BCs).
To render the problem tractable, essential assumptions are applied, leading to a \emph{physics bias}.
Parameters derived or adjusted based on experimental data frequently show \emph{model uncertainties} due to various sources, which then propagate through models across different length and time scales, or fidelities.
These elements introduce significant challenges or `\emph{complexity curses}' in accurately modeling pattern development.

Rapid advancements in machine learning (ML) open new avenues to address these challenges from a data science perspective.
Artificial intelligence (AI) models leveraging neural networks have achieved notable accuracy in recognizing and predicting actions within image sequences or videos~\cite{shi2015convlstm,wang2018predrnn++,gao2022simvp,wang2024ICML}.
Artificial neural networks (ANN) models demonstrated enhanced accuracy and efficiency over conventional numerical solvers~\cite{yang2021Patterns}.
Spiking neural networks (SNNs) are composed of neurons replicating biological activity, which remain dormant until the membrane potential exceeds a threshold.
This feature results in reduced energy consumption compared to the exhaustive approaches seen in ANNs (\zfig{Figs. 1d and 1e})~\cite{maass1997snn,brette2005lif}.
It is also acknowledged that SNNs are better equipped to process information driven by events, which corresponds well to dynamic interactions along with microstructure evolution (\zfig{Fig. 1e}).

The shift from the fixed processing nature of ANNs to the dynamic behavior of SNNs is analogous to the difference between pixel and vector graphics with full-field and reduced-dimension representations of spatial patterns, respectively.
Compared to the structured pixel or voxel data where discrete units may not adapt efficiently to different scales or contexts, vector graphics efficiently capture essential features via geometrical forms defined by mathematical equations, thus improving both scalability and adaptability.
Consequently, while ANNs might excel in highly structured environments, SNNs, which dynamically adjust and process information, may provide a more versatile framework for dealing with complex, evolving systems, akin to microstructure dynamics.
Nonetheless, this intuitive proposal has yet to undergo testing.

In this work, we use both ANNs and SNNs to tackle the spatiotemporal complexity in microstructure evolution. 
In this context, utilizing AI models with digital libraries derived from numerical models serves as a testbed or proxy for experimental data, which can be acquired, albeit at a significantly greater expense.
The performance of neural network performance, tested on simulated fatigue crack growth (FCG) and Turing patterns, shows that SNNs can surpass the accuracy-cost trade-off, achieving higher accuracy with fewer parameters and lower memory usage than state-of-the-art (SOTA) ANN models.
Network architecture analysis explains the superior predictive ability for interface evolution in SNN, particularly beneficial for tasks with strong memory constraints.

\section{Methods}
\subsection{Microstructure Evolution}

Microstructural evolution in materials varies with context to form different interface types.
Two representative examples with localized and collective patterns are considered here for their contrasting spatiotemporal complexity.
FCG has sharply defined interfaces affecting structural integrity, while reaction-diffusion systems feature diffuse interfaces important for chemical and biological development.
This study assesses neural network models addressing these complexities, aiming to improve prediction and optimize material performance beyond traditional solvers.

\subsubsection{Fatigue Crack Growth}

Structural health monitoring (SHM) addresses long-term safety and reliability of engineering systems.
Implementing SHM requires low energy consumption and high precision for real-time, on-site state identification and prediction.
Actual fatigue load spectrum often leads to intricate FCG patterns, despite the homogenization of material microstructures within the framework of continuum mechanics.
We introduced a path-slicing technique using the extended finite element method (XFEM) to model these complexities and construct a digital library of FCG~\cite{zhao2023TAML}.

Plates and shells are key structural elements in aerospace engineering, especially for aircraft and spacecraft.
These 2D models allow for preliminary fatigue assessments, validated by industrial practices as recorded in NASGRO~\cite{mettu1999nasgro}.
We used a 2D plate with dimensions of $10$ mm width and $20$ mm height for the FCG library.
An initial $1$-mm long edge crack was created at mid-height.
The plate width was divided into $N_{\rm s} = 7$ segments for path-slicing.
As the crack extends through segments, we sampled tension and shear loads at top and bottom surfaces, using Gaussian load distributions to model realistic temporal patterns.
Linear elastic fracture mechanics (LEFM) evaluates stress and strain, with stress intensity factors (SIFs) obtained from the interaction integral method~\cite{Shih1988JAM}.

The Paris-Erdogan equation describes the relationship between the crack growth rate, ${\rm d}a/{\rm d}N$, and SIF range within the loading cycle, $\Delta K$, in FCG~\cite{paris1963Paris},

\begin{equation}\label{eq1}
	{\rm d}a/{\rm d}N = C(\Delta K)^m,
\end{equation}

\noindent where $C = 9.7\times10^{-12}$ and $m = 3.0$ are material-specific coefficients fitted from experimental data for typical nickel alloys~\cite{park2022C_m_para}.
The deflection of fatigue crack along the path of propagation was determined by using the maximum shear stress criterion (MSC)~\cite{zhang2016crack},
\begin{equation}\label{eq2}
	\theta = \arccos{\frac{3K_{\rm \uppercase\expandafter{\romannumeral2}}^2+\sqrt{K_{\rm \uppercase\expandafter{\romannumeral1}}^4+8K_{\rm \uppercase\expandafter{\romannumeral1}}^2K_{\rm \uppercase\expandafter{\romannumeral2}}^2}}{K_{\rm \uppercase\expandafter{\romannumeral1}}^2+9K_{\rm \uppercase\expandafter{\romannumeral2}}^2}},
\end{equation}
where $K_{\rm \uppercase\expandafter{\romannumeral1}}$ and $K_{\rm \uppercase\expandafter{\romannumeral2}}$ are mode-$\rm \uppercase\expandafter{\romannumeral1}$ and $\rm \uppercase\expandafter{\romannumeral2}$ SIFs, respectively.
Finally, a digital library comprising time series images with a pixel size of $0.075$ mm and $8$ time steps was constructed, totalling $908$ samples.
This database is subsequently used to resolve the complexity of microstructure evolution through the application of the ML models (\zfig{Supplementary Note 1}).

\subsubsection{Turing Patterns}

Material microstructures are vital for its performance.
Advanced processing technologies use reaction-diffusion to tailor them by adjusting reaction rate, temperature, and chemical concentration~\cite{shim2015NC}.
This control over grain structures, phase interfaces, and porosity optimizes properties such as mechanical strength, electrical conductivity, and thermal stability, matching application needs.

Turing patterns are crucial in materials engineering and natural phenomena, which model complex structures and offer insights into self-organization in materials and biology.
Engineers can predict and manipulate microstructures to develop advanced materials with specific properties by using, for example, the Gray-Scott equations~\cite{pearson1993Science},

\begin{equation}\label{eq3}
	\frac{\partial u}{\partial t} = D_{u}\nabla ^2u -uv^2 +f(1-u),
\end{equation}
\begin{equation}\label{eq4}
	\frac{\partial v}{\partial t} = D_{v}\nabla ^2v +uv^2 -(f+k)v,
\end{equation}

\noindent where $u(x, y)$ and $v(x, y)$ are the concentration fields of chemical species ($U$, $V$) in a 2D space $(x, y)$.
$D_{u}$ and $D_{v}$ are the diffusivity, $f$ is the feed rate of $U$, and $k$ is the rate at which $V$ is removed from the system.
We set parameters $D_{u} = 0.12, D_{v} = 0.08, f = 0.02, k = 0.05$ to obtain Turing patterns with diffusive interface characteristics~\cite{pearson1993Science}.
By randomizing the initial conditions and sampling using sliding windows, we constructed a digital library of Turing patterns containing $720$ samples with a pixel size of $1$ mm and $20$ time steps (\zfig{Supplementary Note 1}).

\subsection{Machine Learning Models}
ML models have been widely used for spatiotemporal predictive learning with datasets like Moving MNIST and KTH actions~\cite{shi2015convlstm,wang2018predrnn++,gao2022simvp,wang2024ICML,srivastava2015fclstm,tan2023openstl}.
The models, classified as ANN and SNN based on neuronal signal processing, assess microstructure evolution complexity (\zfig{Figs. 2a and 2b}).
Their architectures, hyperparameters, data splits, and protocols implemented are detailed in \zfig{Supplementary Note 1}.

\subsubsection{Artificial Neural Network Models}

ANNs can be mathematically formulated as
\begin{equation}\label{eq5}
	\mathbf{o}_n = \phi(\mathbf{W}_n\mathbf{o}_{n-1}+\mathbf{b}_{n}),
\end{equation}
where $\mathbf{o}_{n-1}$, $\mathbf{o}_{n}$, $\mathbf{W}_n$, and $\mathbf{b}_{n}$ are input and output activations, synaptic weight, and bias, respectively.
$n$ is the layer index.
$\phi$ is a nonlinear activation function.
Our base ANN framework combines convolutional neural networks (CNNs) with either recurrent neural networks (RNN) or long short-term memory (LSTM) networks (\zfig{Fig. 2a})~\cite{srivastava2015fclstm,lecun1998cnn,hochreiter1997lstm}.
In this architecture, a CNN-based encoder extracts the spatial patterns, while LSTM networks are employed for the temporal dynamics.
A decoder leveraging CNNs then predicts microstructure evolution from the encoded features.
In addition to address spatial and temporal characteristics separately, we also employ ConvLSTM (\zfig{Fig. 2c})~\cite{shi2015convlstm} that simultaneously capture spatiotemporal data, advanced neural frameworks such as PredRNN++ with a spatiotemporal LSTM (ST-LSTM) cell (\zfig{Fig. 2d})~\cite{wang2018predrnn++}, and SOTA CNN models tailored for video prediction, such as SimVP (\zfig{Fig. 2e})~\cite{gao2022simvp}.

\subsubsection{Spiking Neural Network Models}
Spiking neurons emulate the functions of biological neurons for signal processing, in contrast to artificial neurons.
SNNs can be implemented as~\cite{he2020NeuralNet}
\begin{equation}\label{eq6}
	\mathbf{u}_n^t = e^{-\frac{{\rm d}t}{\tau}}\mathbf{u}_n^{t-1}\odot (1-\mathbf{o}_n^{t-1})+\mathbf{W}_n\mathbf{o}_{n-1}^t,
\end{equation}
\begin{equation}\label{eq7}
	\mathbf{o}_n^t = f(\mathbf{u}_n^t-\mathbf{u}_{\rm th}),
\end{equation}
where $t$ is the timestep index.
$e^{-\frac{{\rm d}t}{\tau}}$ represents the leakage effect of the membrane potential.
$\odot$ denotes the Hadamard product, and $f$ is the step function.
$\mathbf{u}_{\rm th}$ is the firing threshold.
Event-driven signal processing significantly cuts down computational expenses by integrating the membrane potential concept from biological neurons, thereby providing memory functions.
A base encoder-decoder architecture was constructed by integrating CNNs with SNNs (\zfig{Fig. 2b}).
We then implemented the spatiotemporal circuit leaky integrate-and-fire (STCLIF) model that includes autaptic synaptic circuits to capture complex evolutionary patterns at the interfaces (\zfig{Fig. 2f})~\cite{wang2024ICML}.

The rationale for employing models with varying degrees of complexity was twofold, that is, to to enable an extensive assessment of microstructural spatiotemporal predictions across various models, and to ensure that the chosen model complexity suitably aligns with the detailed features inherent in the microstructure data, which will be detailed in the Discussion section.

\section{Results}

\subsection{Data Representations}
Microstructure patterns are often described using field data (e.g., $p(\mathbf{r}, t)$ or $\mathbf{p}(\mathbf{r}, t)$), where fundamental physics such as particle trajectories and interface transformations are embedded in empirical evolutionary models of $p$ or $\mathbf{p}$ (\zfig{Fig. 1}).
Reformulating data into discrete events enhances the detailing of these processes (\zfig{Fig. 1c}).
Instead of relying on the equations of evolution and BCs ascribed by, e.g., the Paris-Erdogan law or reaction-diffusion equations, we construct digital libraries of time series images and leverage ML models to capture the microstructural complexity.
This approach incorporates all spatiotemporal details into the digital library of numerical or experimental data if the fidelity is assured, allowing us to address fundamental physics without the rigid constraints of model assumptions and uncertainties in parametrization.
The same data representation in the digital libraries are used in the ANNs and SNNs studies.

Although ANN frameworks offer higher accuracy in assessing continuous (pixel) data in computer vision (CV), they require more computational resources (\zfig{Fig. 1d})~\cite{shi2015convlstm,wang2018predrnn++,gao2022simvp}.
SNN-based models, unlike ANNs, process event-driven spike signals, reducing computational costs but often sacrificing accuracy, leading to an accuracy-cost trade-off in typical CV tasks (\zfig{Fig. 1e})~\cite{tavanaei2019NeuralNet}.
The unique interface characteristics in microstructural evolution align with the event-driven nature of SNN, positioning it to address this trade-off and tackle the spatiotemporal complexity.

To measure the performance of neural networks, the accuracy of the prediction for an FCG crack or a Turing pattern can be evaluated using the mean absolute error (MAE),
\begin{equation}\label{eq8}
	{\rm MAE} = \sum_{i=1}^{N_x} \sum_{j=1}^{N_y}\frac{|{\rm P}(i,j)-{\rm GT}(i,j)|}{N_xN_y},
\end{equation}
where $\rm P$ and $\rm GT$ are the prediction and ground truth of the microstructures, respectively.
$N_x$, $N_y$ are the discrete dimensions of the region of interest along $x$ and $y$ directions.
For the FCG and Turing examples, the values are $N_x, N_y = 132, 96$ and $200, 200$, respectively.

On the other hand, the estimation of memory utilization can be estimated from data representation and neural network parametrization.
To represent data in pixels, the memory cost ($M$) can be estimated from the data type, $S$(dtype), and the number of grids, $N_{\rm g} = N_x \times N_y$, as $M_{\rm p} = S({\rm dtype}) \times N_{\rm g}$.
In mathematical representation of the geometry, e.g., a curve by nodes on its path, the memory cost is $M_{\rm v} = S({\rm nodes}) \times N_{\rm n} +T = 2 \times S({\rm dtype}) \times N_{\rm n} + T$, where $S({\rm nodes}) = 2 \times S({\rm float}32)$ is the memory used by nodes, each with $2$ components ($x$, $y$).
$N_{\rm n}$ is the number of nodes in discrete paths, which is much smaller than $N_{\rm g}$.
$T$ is the memory used by the transformation matrix ($2\times3$, $4$ bits for each element).
For instance, the pixel-based representation of a typical fatigue crack requires $S({\rm float}32) \times N_x \times N_y = 47.44$ kB, while the vector-based one needs only $2\times S({\rm float}32) \times 40 +24 = 0.344$ kB.
Thin interfaces with continuity in pixels and few-pixel thickness can be represented in vector graphics.
In addition, from the architecture perspective, we also estimate the memory cost of artificial and spiking neurons.
In neural networks, memory cost is mainly determined by the number of parameters, $M_{\rm m} = S({\rm dtype}) \times N_{\rm p}$.
Spiking neurons save about $3/4$ of memory compared to artificial neurons in LSTM, as LSTM cells need $4$ times the parameters for input, forget, update, and output gates, while spiking neurons process temporal data using physiological models~\cite{hochreiter1997lstm,maass1997snn}.

\subsection{Predicting Microstructure Evolution}
The evolution of microstructure patterns in physical systems can be characterized by their varying-interface features (see \zfig{Methods} and \zfig{Fig. 3a} for more information).
We examined the prediction accuracy by analyzing the problems of FCG with a distinct interface and Turing patterns with blurred interfaces.
In the FCG example, a single interface or crack in evolves under complicate loading conditions.
We start from base models such as ANNs (RNN, LSTM) and SNNs to learn and predict pattern development (see \zfig{Methods} and \zfig{Figs. 2a and 2b}).
The findings show that the predictions generated by ANN/LSTM and SNN models closely align with the actual data, while ANN/RNN models even fail to predict with accuracy (\zfig{Figs. 3b and 3c}).
Specifically, crack curve predicted by base ANN/RNN exhibits undesirable discontinuity, whereas the cracks predicted by base ANN/LSTM and SNN appear continuous.
The crack thicknesses, measured in terms of pixels, are $8$ for base ANN/LSTM and $1$ for base SNN, respectively, highlighting the significance of SNN in reducing the dimension of data compared to the full-field pixel representation.

For the Turing patterns, which entail the interaction and evolution of multiple interfaces, base models such as ANN/RNN, ANN/LSTM, and SNN are all inadequate (\zfig{Figs. 3d and 3e}).
ANN/LSTM, in particular, generates a blurred prediction with evenly distributed pixel values, inadequately representing the interface features of microstructures (\zfig{Fig. 3e}).
Conversely, while SNN more accurately depicts the interface features compared to ANN/LSTM, the localization of the interface is still vague and misaligned with the ground truth (\zfig{Fig. 3e}).
These findings indicate that ML models must be advanced in complexity to accurately capture the detailed characteristics of microstructure data with collective spatiotemporal patterns.
Consequently, we utilize advanced spatiotemporal predictive models like ConvLSTM, PredRNN++, SimVP, and STCLIF (see \zfig{Methods} and \zfig{Figs. 2c-f}).
Specifically, the continuity of interfaces predicted by these advanced spatiotemporal predictive models is much improved compared to the base ANN and SNN models.
The interfaces predicted by STCLIF and SimVP have a minimal interface thickness of $2$, whereas those predicted by ConvLSTM, PredRNN++, and the base SNN have an interface thickness of $8$.
Furthermore, STCLIF predicts the interface position more accurately than ConvLSTM, PredRNN++, and SimVP (\zfig{Fig. 3f}).
The results indicate that the STCLIF model outperforms ANN-based counterparts in prediction accuracy (\zfig{Fig. 3f}).

\subsection{Quantitative Model Evaluation}
Quantitative evaluation indicates that with longer observation duration, the predictive error concerning crack morphology in the FCG problem decreases in both base ANN/LSTM and base SNN.
This fact is attributed to the larger dataset available over extended observation times (\zfig{Fig. 4a}).
Conversely, the accuracy of base ANN/RNN show no improvement for its simpler architecture and fewer parameters compared to ANN/LSTM, which limit its capability to capture complex spatiotemporal patterns~\cite{jozefowicz2015ICML}.
Compared to base ANNs (RNN, LSTM), the base SNN model demonstrates a substantial reduction in the number of parameters, by $4-5$ orders of magnitude, while preserving prediction accuracy (\zfig{Fig. 4b}).

Regarding Turing pattern prediction, the evaluation of quantitative outcomes measures the MAE between predicted and actual patterns across various time steps.
The STCLIF model registers the least predictive and cumulative errors, achieving a $27.71\%$ error reduction compared to the ANN models (\zfig{Fig. 4c}).
Importantly, the STCLIF presents an optimal balance of accuracy and cost in the context of Turing pattern development (\zfig{Fig. 4d}). 
Specifically, it surpasses the ANN model in accuracy while reducing the number of parameters by an order of magnitude (\zfig{Fig. 4d}).

FCG and Turing patterns exhibit contrasting, localized versus collective, spatiotemporal complexities, and this difference can be reflected in the quantitative evaluation of ML models.
For ML models that effectively capture the spatiotemporal complexities of FCG and Turing patterns (base SNN and STCLIF with MAE < $0.01$), the parameter count shows that only $1,000$ parameters are sufficient to effectively model the spatiotemporal complexity of FCG (base SNN, \zfig{Fig. 4b}), whereas $3$ million parameters are required to model the spatiotemporal complexity of Turing patterns (STCLIF, \zfig{Fig. 4d}).
This result highlights the link between the spatiotemporal complexity of data and model complexity.

\subsection{Network Architecture Analysis}
The investigation into the architecture of well-trained neural networks shows that the STCLIF surpasses ConvLSTM in managing the complexity of microstructure evolution.
Examination of the weight distribution of convolutional layer ($\mathbf{W}_n$ in \zfig{Eq. 5} and \zfig{Eq. 6}) demonstrates that STCLIF features weights with reduced variance and smaller magnitudes than those in ConvLSTM (\zfig{Fig. 5a}).
This feature usually indicates better generalization, similar to regularization techniques in ML, in which a penalty term ($\lambda \Sigma w^2$, where $w$ are the elements of matrix $\mathbf{W}_n$) is added to the loss function, encouraging simpler models that generalize better to unseen data~\cite{krogh1991NIPS}.
We apply a small weight threshold to consider neurons unlinked if the weight is below this level ($|w|<0.001$), which helps evaluate the network connectivity density.
The STCLIF model shows significantly lower connectivity density compared to the ConvLSTM ($22.07\%$), often linked to improved generalization akin to dropout techniques in ML, in which neurons are dropped with a certain probability at each iteration during training to prevent overfitting and improve model generalization~\cite{srivastava2014dropout} (\zfig{Fig. 5b}).
This feature can be explained by the fact that spiking neurons utilize biologically-inspired temporal dynamics and event-driven computation, unlike the continuous operations of artificial neurons.
This results in sparser connectivity, activating neurons only at threshold potentials, which enhances sparsity.
Consequently, the STCLIF is more effective and efficient than ANNs in managing complex microstructural evolution driven by interfacial dynamics.

\section{Discussion}
\subsection{Related Works}
Spatiotemporal neural networks such as ANNs and SNNs extract features for predictive tasks.
ConvLSTM integrated CNN with LSTM, using convolutional in place of fully connected operations for modeling~\cite{shi2015convlstm}.
PredRNN and PredRNN++ improved on ConvLSTM by introducing ST-LSTM cells with additional memory cells to enhance memory flow between layers~\cite{wang2018predrnn++}.
Despite advancements, these recurrent models face parallelization challenges. Recurrent-free models like SimVP use CNNs in an encoder-decoder architecture to convolve spatiotemporal channels, achieving SOTA performance on Moving MNIST datasets.
Transformer-based networks such as Vision Transformer and TimeSformer focus on video classification but have limited use in spatiotemporal prediction~\cite{dosovitskiy2020ViT,bertasius2021ICML,tang2024predformer}.
Large video generation models like SORA and Kling use diffusion transformers for pixel-based videos~\cite{peebles2023CVPR,openai2024sora,kuaishou2024kling}.
However, they face scalability issues due to high computational and memory demands.
SNNs with leaky integrate-and-fire (LIF) neurons perform less accurately in capturing spatiotemporal features in CV compared to ANNs on datasets like Moving MNIST and KTH~\cite{safa2021convsnn,zhang2024tclif,tavanaei2019NeuralNet}.
The STCLIF model, inspired by autaptic synapses, introduces spatiotemporal self-connections to enhance feature extraction and accuracy, yet it still lags behind leading ANN models in CV tasks~\cite{wang2024ICML}.

ANN models have shown potential in learning and predicting microstructure evolution, enhancing computational efficiency over traditional simulations~\cite{yang2021Patterns,alhada2024npjcm}.
On the other hand, vector-like geometrical abstraction reduces memory usage, enhancing efficiency in large-scale, real-time engineering applications like SHM, where energy efficiency is critical.
SNNs were utilized in the nonlinear regression of extensive sensor data by averaging the membrane potentials of neurons, producing real-valued outputs to evaluate mechanical stress and deformation of materials~\cite{henkes2024RSOS}.
Our framework based on STCLIF thus supports digital twin-based SHM for energy-efficient, on-site fatigue crack prediction and real-time material processing control.
However, no study has yet utilized the event-driven capabilities of SNN models to explore microstructure evolution complexity.

Numerous ML models such as physics-informed neural networks (PINNs), Kolmogorov-Arnold network (KAN), and operator learning (deep operator network or DeepONet, spiking DeepONet)~\cite{raissi2019PINN,liu2024kan,wang2025CMAME,lu2021deeponet,yang2023PNAS,kahana2022PR}, have been deployed to solve PDEs while imposing strong mathematical constraints, thereby improving efficiency and generalization under the constraint.
Challenges arise in dealing with phenomena involving multiple physics and scales, or problems lacking governing PDEs. 
To tackle this problem, we utilize a `digital library' approach that applies weak physics constraints on neural networks compared to the mathematical confines prescribed by the PDEs.
This allows us to explore the complexity of microstructure evolution and comprehend network functions after the fact~\cite{zhao2024PR}.
The digital libraries can be constructed from high-fidelity simulations, high-resolution experiments, or generative models with essential physics~\cite{ohana2024well}. 

Interpreting neural networks involves visualizing components like weights and feature maps to understand information flow~\cite{simonyan2013PR}.
Techniques such as Shapley additive explanations (SHAP) and symbolic regression clarify model input contributions~\cite{lundberg2017SHAP,koza1994SR}.
Despite extensive efforts by researchers, the interpretability of ML models remains a significant challenge~\cite{messeri2024Nature}.
Our method relies on the digital library of physical principles, which effectively handles complexities beyond PDEs, making it suitable for integration with ML techniques and analysis of physical phenomena.
The spatiotemporal complexity in digital libraries is reflected in the neural network architecture, and analyzing the architecture can help explore physics from first principles while mitigating biases and uncertainties.

\subsection{Resolving Data and Model Complexity}
Simplifying high-dimensional data is essential for understanding microstructural evolution across spatial and temporal dimensions.
Biologically inspired SNNs such as STCLIF leverage intrinsic dynamics and sparsity to to simplify model representations.
They align better with microstructure interfaces than ANNs, offering more effective simplification of data complexity.

Pattern evolution in physical systems falls into two main scenarios with distinct modeling approaches.
The LS method is used for problems where the interface is important but the bulk phase is not. PF models are employed when both interface and bulk phase details are crucial.
ANNs process spatial characteristics through static, structured behaviors, such as layers of networks and local connections between neurons.
While traditional ANNs do not explicitly handle temporal characteristics, architectures like RNN and LSTM are designed to model temporal characteristics by retaining memory of past states.
SNN models including base SNN and STCLIF, on the other hand, model spatiotemporal characteristics through the coupling of time and space, utilizing the timing of spikes, temporal encoding, and the synaptic connections, offering a closer analogy to biological systems.
STCLIF excels in interface evolution challenges, delivering high accuracy and efficiency~\cite{hughes2003FEM}.
In practice, SNN models with suitable complexities should be used as exemplified in FCG and reaction-diffusion systems, where STCLIF outperforms the base SNN model, which perform well only in simple FCG scenarios.
ANNs, while more accurate for bulk phase-focused problems and those with multiphysics field coupling, require higher computational resources.
Selecting appropriate SNN or ANN models is essential for predicting evolutionary data in experiments, considering interface, bulk phase, and problem complexity.
Evaluating ML model performance helps in choosing the right method for modeling experimental microstructure evolution.
In the digital library approach, superior SNN performance over ANN indicates that the LS method manages the complexity well.
The challenge of energy consumption is particularly significant in embedded systems, notably in aerospace applications and embodied intelligence.
Our SNN framework based on the STCLIF displays high accuracy and low computational cost for microstructure evolution analysis, making it perfect for digital twin scenarios.

\subsection{Scaling with dimensions}
Although not explored in this work, scaling from $2$D to $3$D shifts data representation from $2$D lines to $3$D surfaces, moving from pixel to voxel data.
For FCG cracks or Turing patterns, memory overhead increases, proportional to discretization due to the added dimension ($100-1000$ times).
Geometries defined by mathematical equations via significantly fewer data points in the reduced-dimension representation, from curves in 2D to surfaces in 3D, significantly reduces memory costs compared to voxel data.
ML models for $3$D microstructure evolution need $2$D to $3$D convolutions, raising memory usage for parameters with an increase proportional to the convolution kernel size ($3-7$ times).
The memory data overhead is substantial, but the increase in model parameters is smaller.
The SNN model with $3$D convolutions saves about $3/4$ of memory compared to ANN combining $3$D convolutions and LSTM.
Limitations in experimental digital libraries include collecting high-quality 3D data and boosting model robustness against measurement noise through noise-inclusive training.

\section{Conclusions}
We use artificial and spiking neural networks to tackle the spatiotemporal complexity of microstructure evolution in matter, exemplified by fatigue crack propagation and Turing pattern development.
Both of the two representative problems hold considerable importance in engineering contexts, such as structural health monitoring and material processing control.
Our study shows that spiking neural networks balance accuracy and computational cost in spatiotemporal tasks for microstructure evolution, in contrast to the common practices in computer vision tasks.
Specifically, compared to artificial neural networks, spiking neural networks such as the spatiotemporal circuit leaky integrate-and-fire model match the accuracy with fewer parameters and less memory usage due to reduced weights and sparser connections.
Moreover, the digital library approach mitigates the strong mathematical constraints to the physics of problems under investigation, and can be extended to other problems such as grain growth and phase transformations.
These findings and discussions guide the design of machine learning models for microstructure evolution.

\clearpage
\newpage

\section*{Declaration of Competing Interest}

\noindent The authors declare that they have no competing financial interests.

\section*{Credit authorship contribution statement}
Z.X. conceived the concept.
Y.Z. developed the methodology and performed the computational study.
Both authors analyzed the data and wrote the manuscript.


\section*{Acknowledgements}
\noindent This work was supported by the National Natural Science Foundation of China through grants 12425201 and 52090032.
The computation was performed on the Explorer $1000$ cluster system of the Tsinghua National Laboratory for Information Science and Technology.

\clearpage
\newpage

\bibliographystyle{iopart-num.bst}
\bibliography{main}

\clearpage
\newpage

\begin{figure}[H]
\centering
\includegraphics[width=\linewidth] {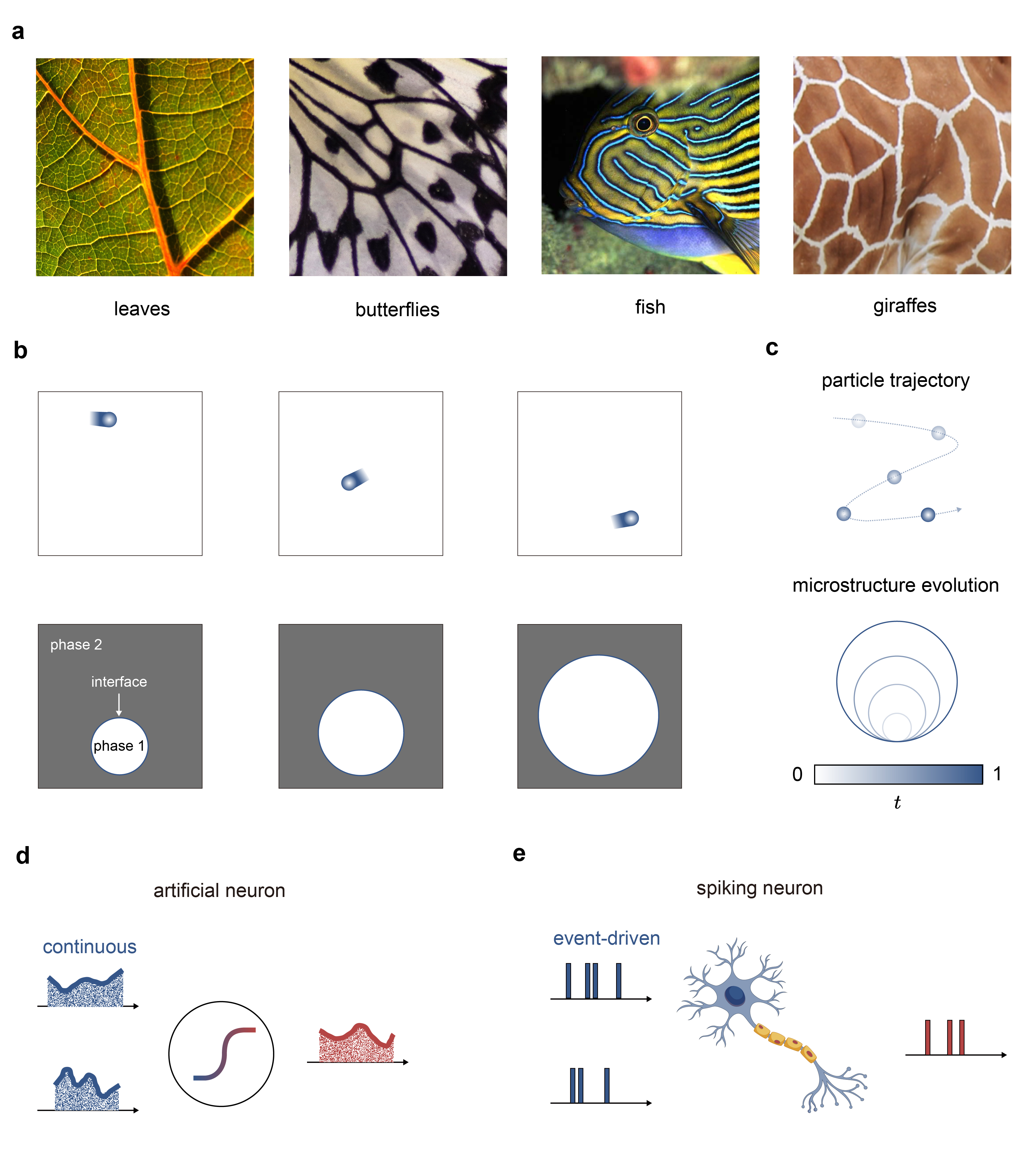} 
\caption{{\bf Representations of spatiotemporal complexity.}
{\bf (a)} Turing patterns in nature.
{\bf (b)} Evolution of physical systems in pixel/voxel representations, encompassing information not necessary for the governing physical laws.
{\bf (c)} Interfaces play a crucial role in characterizing the development of physical systems.
{\bf (d)} Artificial neurons, which can be used to process continuously evolving pixel/voxel information.
{\bf (e)} Spiking neurons, capable of processing information driven by event-based spikes.
}
\label {fig1}
\end{figure}

\clearpage
\newpage

\begin{figure}[H]
\centering
\includegraphics[width=\linewidth] {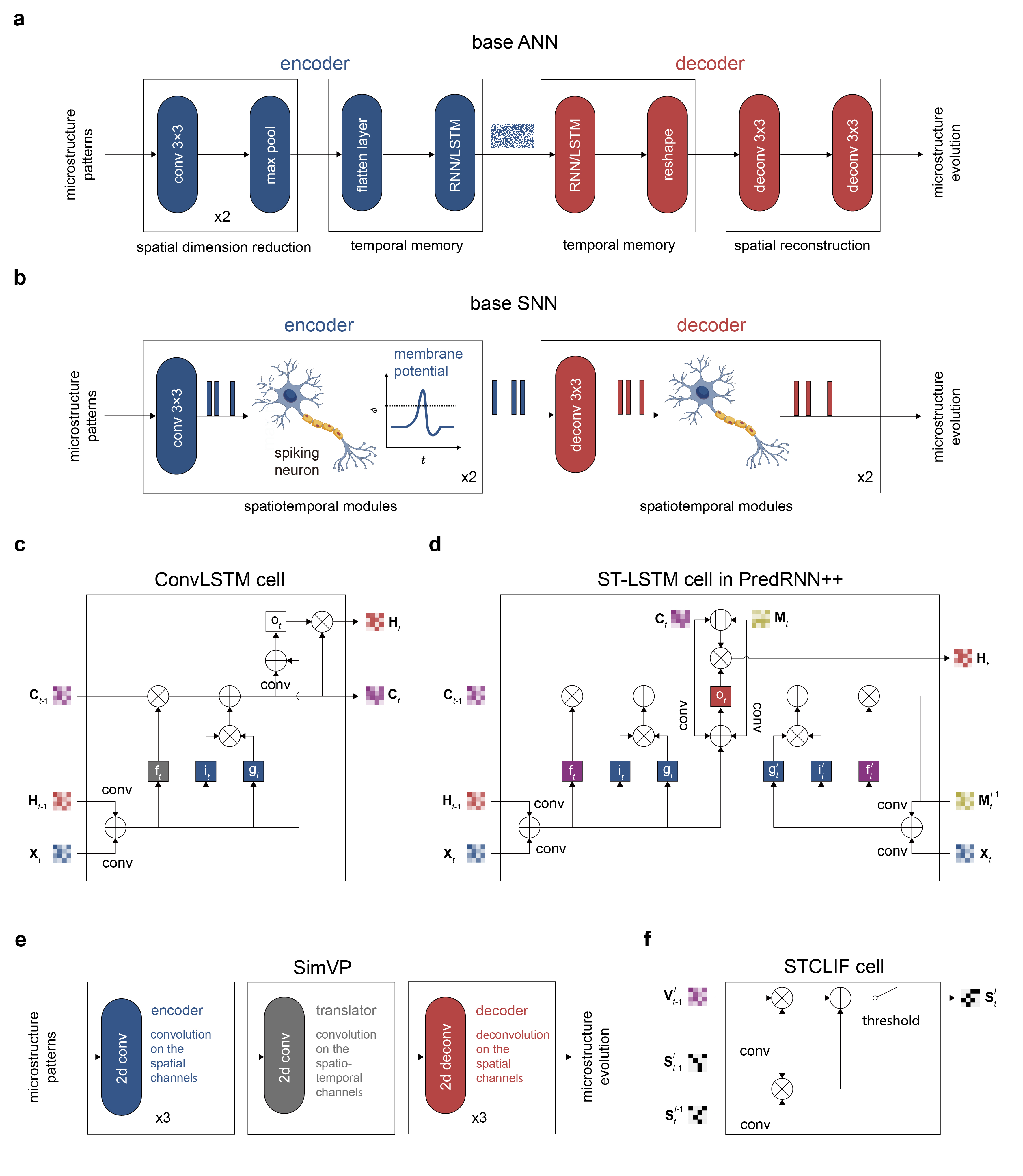} 
\caption{{\bf Neural network prediction using artificial and spiking neurons.}
{\bf (a)} Neural networks separately process spatial and temporal information and combine them subsequently within artificial neural networks (ANNs).
{\bf (b)} In spiking neural networks (SNNs), spatiotemporal data is processed in an integrated manner by neural networks.
{\bf (c-f)} The architectures of advanced spatiotemporal predictive models include ConvLSTM {\bf (c)}, PredRNN++ {\bf (d)}, and SimVP {\bf (e)} based on ANNs, as well as STCLIF {\bf (f)} based on SNNs.
${\rm f}_t$, ${\rm i}_t$, ${\rm g}_t$, and ${\rm o}_t$ are forget gate, input gate, input-modulation gate, and output gate, respectively.
${\bf X}$, ${\bf H}$, ${\bf C}$, ${\bf M}$, ${\bf S}$, ${\bf V}$ are inputs, hidden states, temporal cell states, spatiotemporal memory states, spiking input, and membrane potentials, respectively.
Superscripts denote the layer index, while subscripts represent the time step.
The symbols $\otimes$, $\oplus$, and $\parallel$ represent the Hadamard product, pointwise addition, and concatenation operation, respectively.
}
\label {fig2}
\end{figure}

\clearpage
\newpage

\begin{figure}[H]
\centering
\includegraphics[width=\linewidth] {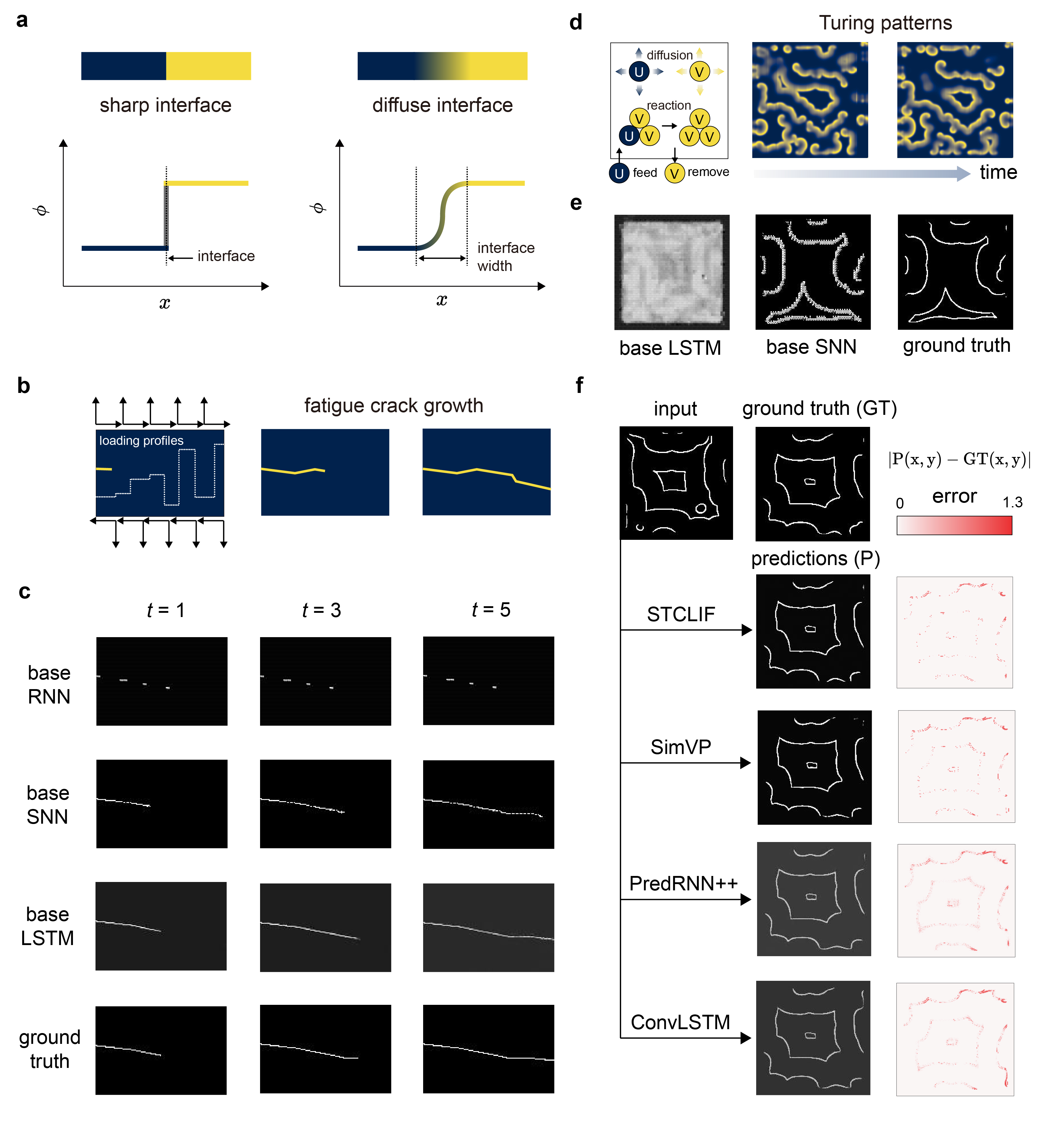} 
\caption{{\bf Interface representations in microstructure evolution.}
{\bf (a)} Sharp and diffuse interfaces in modeling microstructure evolution.
{\bf (b)} Fatigue crack growth (FCG) with sharp interfaces.
{\bf (c)} FCG prediction using base ANN (RNN and LSTM) and SNN models.
{\bf (d)} Evolution of Turing patterns with diffuse interfaces.
{\bf (e)} Limitations of base ANN/LSTM and SNN models in predicting Turing patterns.
{\bf (f)} Predicted microstructure evolution for Turing patterns using advanced spatiotemporal predictive models.
}
\label {fig3}
\end{figure}

\clearpage
\newpage

\begin{figure}[H]
\centering
\includegraphics[width=\linewidth] {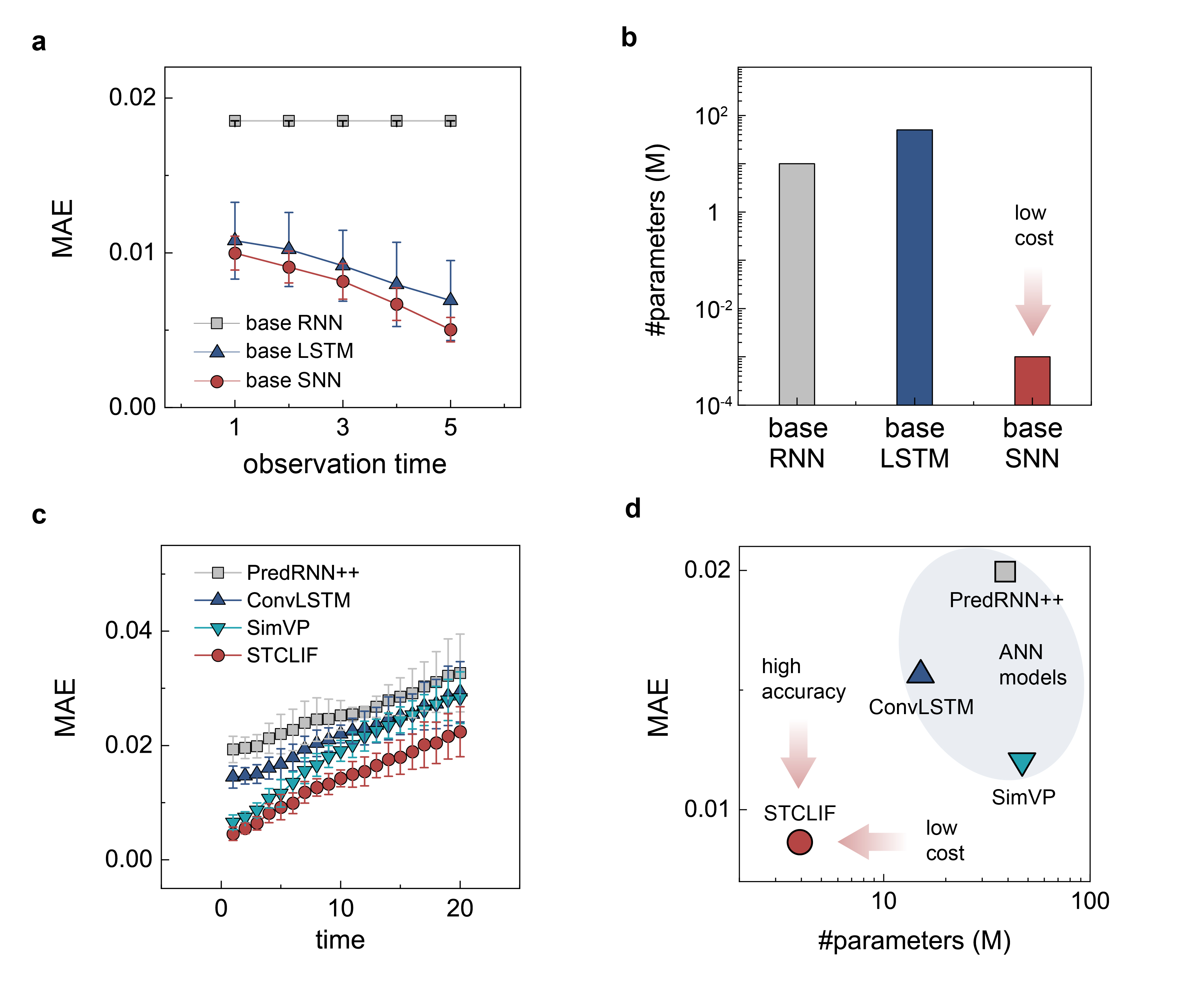} 
\caption{{\bf Accuracy and cost of microstructure evolution prediction.}
{\bf (a)} Mean absolute error (MAE) of FCG prediction and ground truth of the crack morphologies (\zfig{Eq. 8}).
{\bf (b)} The number of parameters used in the neural network model for FCG prediction.
{\bf (c)} MAE of Turing patterns prediction and ground truth of the microstructures (\zfig{Eq. 8}).
{\bf (d)} The accuracy-cost map of advanced spatiotemporal models (ConvLSTM, PredRNN++, SimVP, STCLIF) in predicting Turing patterns.
}
\label {fig4}
\end{figure}

\clearpage
\newpage

\begin{figure}[H]
\centering
\includegraphics[width=\linewidth] {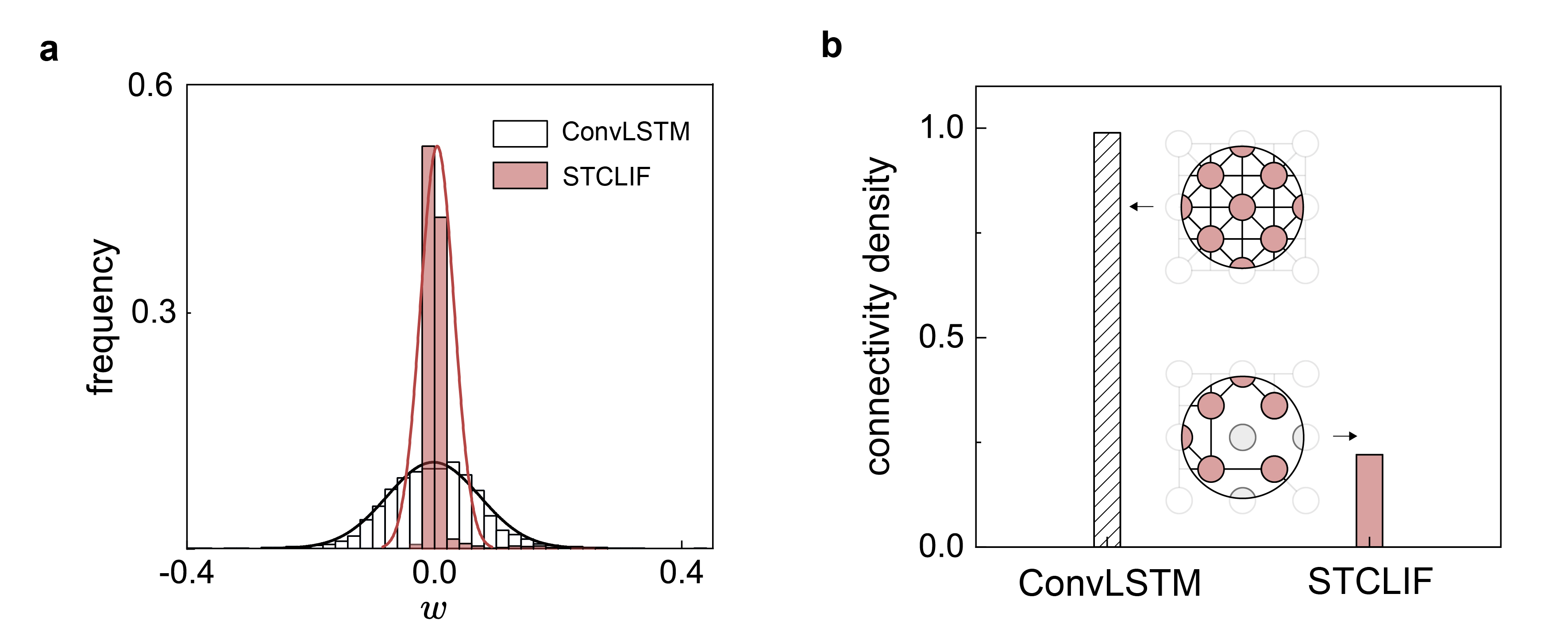} 
\caption{{\bf Neural network architectures.}
{\bf (a)} Weights distribution of convolutional layers in ConvLSTM and STCLIF models.
{\bf (b)} Connectivity patterns in ConvLSTM and STCLIF models.
}
\label {fig5}
\end{figure}

\clearpage
\newpage





\section*{Supplementary Information Material}
\noindent This Supplementary Information Material includes Supplementary Note 1.

\clearpage
\newpage



\setlength{\parskip}{6 pt}

\section*{Supplementary Note 1. Details of machine learning (ML) models}

\subsection*{Neural network architectures and hyperparameter settings}
To make a comprehensive comparison for predicting microstructure evolution and apply it to microstructure data of varying complexity, we developed base machine larning (ML) models (RNN, LSTM, and SNN) (\zfig{Fig. 2}) and employed spatiotemporal predictive learning models (ConvLSTM, PredRNN++, SimVP, and STCLIF)~\cite{shi2015convlstm,wang2018predrnn++,gao2022simvp,wang2024ICML}.

For base ANN models, an encoder-decoder architecture is adopted to extract the spatiotemporal features of microstructure evolution.
For the encoder, two $3$D convolutional layers are employed, with channel numbers of $1$ (input of grayscale image), $16$ (latent channels of the first layer), and $4$ (latent channels of the second layer), and a kernel size of $3\times3\times3$.
The extracted spatial features are then converted into a vector ($3,128$ dimensions) using a flattened layer.
Following this, spatial features are further processed using two layers of RNNs or LSTMs, with $128$ and $3,128$ hidden neurons, respectively.
Finally, a decoder consisting of two deconvolution layers is used to predict microstructure evolution based on the spatiotemporal features extracted by the encoder in which the channel numbers are $16$ and $4$, with a kernel size of $3$.

For the base SNN model, a similar encoder-decoder architecture is adopted.
Since spiking neurons have memory characteristics, we use $2$D convolutional and deconvolutional layers without the need for additional flattened layers or RNN/LSTM layers for temporal modeling.
The channel dimensions and kernel sizes are consistent with those used in the RNN/LSTM models.
The LIF neuron model~\cite{brette2005lif} is employed to model spiking neurons, with hyperparameters such as the membrane potential decay rate of $0.5$, and a membrane threshold of $1$.
For more sophisticated spatiotemporal predictive learning models such as ConvLSTM, PredRNN++, SimVP, and STCLIF, refer to the original published literature~\cite {shi2015convlstm,wang2018predrnn++,gao2022simvp,wang2024ICML}.

\subsection*{Input-output configurations, and the setups of training and testing}

For the FCG in a single-crack scenario, the pattern of interface evolution is relatively simple.
Therefore, we use simple models (vanilla RNNs, LSTM, and SNNs) to predict its microstructure evolution.
By sampling load amplitudes from a Gaussian distribution (the mean is $200$ MPa for tension loading and $100$ MPa for shear loading, the variance is $50$ MPa for both), $908$ samples were obtained, each with a time series length of $8$.
Among these, $800$ samples were used for training, and the remaining $108$ samples were used for testing to evaluate the model performance.
A sliding window with fixed length $4$ ($3$ frames as input, $1$ frame as output) and sliding step of $1$ was applied to these samples.
As a result, the final size of the training set is $800\times5=4,000$, and the size of the test set is $108\times5=540$.
The model input consists of three microstructure images, and the output is the next microstructure image corresponding to these three input patterns.
The loss function is the mean squared error (MSE) between the predicted and the true microstructure patterns.
The Adam optimizer is used to train the neural network parameters, with hyperparameters settings: a learning rate of $0.001$, and first- and second-moment decay rates $\beta_0 = 0.9, \beta_1 = 0.999$.
In the testing scenario, in addition to directly evaluating the prediction performance (by inputting the true microstructure at each time step and predicting microstructures in the next frame, \zfig{Fig. 3c}), we also adopt an autoregressive approach, where the model predictions are used as the input for the next prediction step, continuing until the final microstructure pattern is predicted (\zfig{Fig. 4a}).

For Turing patterns, where the dynamics involve interaction and evolution of multiple interfaces, simple models like vanilla RNNs, LSTM, and SNNs are not adequate for predicting the pattern evolution (\zfig{Fig. 3e}).
Therefore, we utilize more sophisticated spatiotemporal predictive learning models such as ConvLSTM, PredRNN++, SimVP, and STCLIF models to learn and predict the Turing patterns. 
By randomizing the initial conditions, $15$ samples were obtained, each with a time series length of $68$.
Among these, $10$ samples were used for training, and the remaining $5$ samples were used for testing to evaluate the model's performance.
A sliding window with fixed length $20$ ($10$ frames as input, $10$ frames as output) and sliding step of $1$ was applied to these samples.
As a result, the final size of the training set is $10\times48=480$, and the size of the test set is $5\times48=240$.
The equations are dimensionless (\zfig{Eq. 3 and Eq. 4}), and the size of pixels is $1$ mm.
The model inputs consist of $10$ microstructure images, and the output is the next $10$ microstructure image corresponding to these $10$ input patterns.
Similar to the settings of the FCG problem, MSE and Adam with a learning rate of $0.001$ and decay rates $\beta_0 = 0.9, \beta_1 = 0.999$ are adopted as the loss function and optimizer, respectively.
The model evaluation during testing also follows an autoregressive approach (\zfig{Fig. 4c}).

\end{document}